\documentclass[11pt]{article}

\usepackage[final]{acl}

\usepackage{times}
\usepackage{latexsym}

\usepackage[T1]{fontenc}

\usepackage[utf8]{inputenc}

\usepackage{microtype}

\usepackage{inconsolata}

\usepackage{graphicx}

\usepackage{amsmath}
\usepackage{amssymb}
\usepackage{mathtools} 
\usepackage{booktabs}
\usepackage{multirow}
\usepackage[table]{xcolor}
\usepackage{array}
\usepackage{colortbl}
\usepackage{subcaption}
\usepackage[most]{tcolorbox}
\usepackage{xcolor}
\definecolor{promptbg}{RGB}{235, 240, 250}

\title{Anchoring What Matters: A Dual-Level Learning Framework for Visually-Grounded Multimodal Reasoning}

\author{
 \textbf{Xinxin Song\textsuperscript{1}\thanks{Equal Contribution.}},
 \textbf{Siyuan Li\textsuperscript{1}\footnotemark[1]},
 \textbf{Tingxiong Xiao\textsuperscript{1}},
 \textbf{Jinli Suo\textsuperscript{1,2}\thanks{Corresponding author.}}
\\
\\
 \textsuperscript{1}Department of Automation, Tsinghua University \\
 \textsuperscript{2}Institute for Brain and Cognitive Science, Tsinghua University
\\
 {
   \texttt{jlsuo@tsinghua.edu.cn}
 }
}

\begin{document}
\maketitle
\begin{abstract}
Reinforcement learning with verifiable rewards (RLVR) has significantly improved the reasoning capabilities of large vision-language models (LVLMs). However, standard on-policy RLVR algorithms face a critical optimization bottleneck in preserving and reinforcing visually grounded reasoning behaviors: valuable visually-grounded reasoning trajectories are discarded after a single update, while uniform token advantage allocation prevents the model from reinforcing critical perception or reasoning steps. To bridge this gap, we propose PIVOT, a dual-level learning framework that anchors policy optimization around informative visual reasoning signals. Specifically, PIVOT introduces a self-calibrated experience replay mechanism, which selectively collects and replays visually-grounded historical experiences as stable reference anchors for policy optimization. Building upon this, we further design a vision-guided advantage allocation mechanism to allocate additional vision-aware advantages to tokens based on their local visual support and impact on downstream reasoning. Extensive experiments across diverse benchmarks demonstrate that PIVOT achieves highly competitive performance in enhancing the multimodal reasoning capabilities of LVLMs.
\end{abstract}

\section{Introduction}

Reinforcement learning with verifiable rewards (RLVR) has become a prevailing post-training paradigm for enhancing the reasoning capabilities of large language models~\citep{zhang2025surveyreinforcementlearninglarge}.  Given its success in the text domain, recent studies have extended RLVR to large vision-language models (LVLMs) and achieved promising results in visual reasoning tasks~\citep{Liu_2025_ICCV}. However, directly applying standard RLVR algorithms to LVLMs exposes a critical optimization bottleneck: sparse yet informative learning signals are neither sufficiently preserved nor effectively exploited, leading to inefficient optimization for visually grounded reasoning. Specifically, this inefficiency manifests at two interconnected levels.

At the \textit{trajectory level}, standard on-policy RL algorithms typically discard the generated trajectories\footnote{We use the terms \textit{trajectory}, \textit{experience}, and \textit{response} interchangeably throughout this paper.} after each update phase. For LVLMs, obtaining high-quality visually grounded trajectories is non-trivial due to the expanded multimodal search space. These trajectories explicitly demonstrate how to ground visual evidence into logical steps, containing richer information than a sparse binary reward. Discarding them indiscriminately results in a waste of optimization signals.
At the \textit{token level}, the issue of uniform credit assignment further dilutes the learning signals. Outcome-based RLVR algorithms typically assign the same scalar advantage to all tokens within a trajectory. However, the visual reasoning chain is highly heterogeneous: certain tokens act as pivotal points for visual perception or logical branching, whereas others are mere linguistic fillers. Assigning uniform optimization strength to all tokens prevents the model from anchoring on the critical perception or reasoning steps, causing informative token-level signals to be diluted across the entire response. 

To address the above limitations, we propose PIVOT, a dual-level learning framework that Preserves and reinforces sparse but Informative learning signals for Visually-grOunded reasoning at both the trajectory and Token levels. Specifically, at the trajectory level, we introduce a self-calibrated experience replay mechanism, which builds a dynamic experience pool to collect and selectively replay informative prompts alongside their high-quality, visually-grounded historical trajectories. These trajectories act as retrospective anchors rather than direct training objectives, gently constraining the policy without stifling exploration. Beyond trajectory-level anchoring, we further introduce a vision-guided advantage allocation strategy. This mechanism obtains fine-grained visual learning signals by calculating the counterfactual visual support of sampled tokens and estimating their impact on future reasoning. These signals are then used for fine-grained advantage allocation to concentrate the optimization strength on more valuable perception and reasoning tokens. The main contributions of our method are summarized as follows:
\begin{itemize}
    \item We identify a key optimization bottleneck in multimodal RLVR: valuable visual reasoning signals are diluted across both trajectory sampling and token-level optimization, making it difficult for the policy to consistently reinforce visually grounded reasoning behaviors.
    \item We propose PIVOT, a unified dual-level framework that preserves high-value visual reasoning experiences as retrospective anchors and reinforces optimization on visually grounded tokens via fine-grained advantage modulation.
    \item Extensive experiments on various data and model scales demonstrate that PIVOT consistently improves multimodal reasoning performance over baselines.
\end{itemize}

\section{Related Works}

\noindent\textbf{RLVR for Multimodal Reasoning.~~~~}
Given the success of RLVR in the text domain, recent works have begun to explore its application to LVLMs. From data perspective, methods like Vision-R1~\citep{huang2026visionr} and MM-EUREKA~\citep{meng2025mmeurekaexploringfrontiersmultimodal} construct high-quality CoT datasets, while NoisyRollout~\citep{liu2025noisyrollout} injects perceptual diversity through data augmentation. For reward design, Perception-R1~\citep{NEURIPS2025_880ee6e2} designs specialized reward mechanisms for different visual tasks, and Vision-SR1~\citep{li2026visionsr} decomposes visual perception and language reasoning to formulate a self-reward mechanism. For optimization objectives, PAPO~\citep{wang2026perceptionaware} introduces an additional KL loss to improve perceptual capabilities. Unlike prior work that mainly improves multimodal RL through stronger supervision signals, we revisit the optimization process and study how high-value visual reasoning signals can be more effectively preserved and exploited during policy learning.

\noindent\textbf{Experience-based RL.~~~~}
Experience replay~\citep{lin1992self} has been extensively studied in classical RL tasks~\citep{9904958}. Recent studies have extended it to LLMs. For instance, RLEP~\citep{zhang2025rlep} introduces a two-stage framework of collection followed by replay; ReMix~\citep{liang2026squeeze} formulates a stable optimization objective for off-policy training; and ExGRPO~\citep{zhan2026exgrpo} prioritizes the replay of high-value trajectories based on systematic metric analysis. For LVLMs, EFRame~\citep{wang2025eframe} replays successful trajectories for hard prompts, while VL-Rethinker~\cite{NEURIPS2025_2c84844a} mitigates the vanishing advantage problem via selective sample replay. In this work, we emphasize visual perception quality during experience collection while treating historical experience as a reference anchor for policy updates rather than a direct optimization objective, which yields better results. CalibRL~\citep{huang2026controllable} adopts a similar optimization objective to ours, but it relies on external expert demonstrations instead of self-generated experience.

\noindent\textbf{Token Optimization in RLVR.~~~~}
Recent works have revealed the heterogeneity and sparsity of the token optimization signal in RLVR updates~\citep{NEURIPS2025_a797c2d2}. In LVLMs, prior works assess token-level visual dependence using counterfactual output divergence~\citep{ye2026not,huang2026spotlight}, hidden state similarities~\citep{wang2026visually}, or visual attention scores~\citep{jiao2026credit,luo2026from} to reinforce tokens with strong perceptual grounding. Methods like PEPO~\citep{li2026rethinking} and ToR~\citep{lu2026bridging} further integrate this with high-entropy reasoning tokens. Building upon counterfactual support for local visual dependence, we further formulate an entropy-gated future visual support to capture downstream impact, enabling fine-grained advantage allocation.

\section{Preliminaries}

\subsection{Multimodal RLVR}
Consider a multimodal reasoning problem where each input $x$ consists of a textual query $q$ and an image $I$, denoted by $x=(q, I)$. An LVLM parameterized by $\pi_\theta$ generates a textual response sequence $\mathbf{y}=(y_1,\ldots,y_T)$ autoregressively according to
\begin{equation}
\pi_\theta(\mathbf{y}\mid x)
=
\prod_{t=1}^{T}
\pi_\theta(y_t\mid q,I,y_{<t}).
\end{equation}
For reasoning tasks, the generated sequence $\mathbf{y}$ typically consists of an intermediate multi-step reasoning chain followed by the final answer. In the RLVR framework, a rule-based verifier assigns a binary reward $R\in\{0,1\}$ to each response based solely on whether its final extracted answer matches the ground truth $a$. 

DAPO~\citep{NEURIPS2025_a4277440} is a widely used RLVR algorithm for reasoning tasks. It samples a group of $G$ candidate responses $\{\mathbf{y}_i\}_{i=1}^G$ from old policy $\pi_{\theta_{\text{old}}}$ and computes the advantage $\hat{A}_i$ for the $i$-th candidate response $\mathbf{y}_i$ by
\begin{equation}
    \hat{A}_i = \frac{R_i - \text{mean}(\{R_k\}_{k=1}^G)}{\text{std}(\{R_k\}_{k=1}^G)}.
\end{equation}
The policy parameters $\theta$ are then updated by minimizing the following clipping objective:
\begin{equation}
\label{eq:grpo_objective}
\begin{aligned}
\mathcal{L}(\theta) &= 
-\mathbb{E}\Bigg[ \frac{1}{\sum_{i=1}^G|\mathbf{y}_i|} \sum_{i=1}^{G} 
\sum_{t=1}^{|\mathbf{y}_i|} 
\min \Big( \rho_{i,t}(\theta) \hat{A}_{i,t}, \\
&\qquad \text{clip}\big(\rho_{i,t}(\theta), 1 - \epsilon_l, 1 + \epsilon_h\big) \hat{A}_{i,t} \Big)
\Bigg],
\end{aligned}
\end{equation}
where $\rho_{i,t}(\theta) = \frac{\pi_\theta(y_{i,t} | q,I, \mathbf{y}_{i,<t})}{\pi_{\theta_{\text{old}}}(y_{i,t} | q,I, \mathbf{y}_{i,<t})}$ is the importance sampling ratio, $\epsilon_l$ and $\epsilon_h$ are the clipping hyperparameters.


\subsection{Trajectory Quality Estimation}
\label{pre:tqs}
To evaluate the quality of generated trajectories in multimodal RLVR, we consider two complementary signals that capture different aspects of trajectory reliability.

We first consider the visual dependency of a trajectory. Following previous work~\citep{huang2026spotlight}, we quantify it by computing the KL divergence between the predictive distribution of the policy conditioned on the original image and a corrupted version. Given the multimodal input $(q, I)$ and the generated trajectory $\mathbf{y}$, we will perform counterfactual interventions (such as random patch masking~\citep{wang2026perceptionaware}) on the original image $I$ to obtain a corrupted image $\widetilde{I}$, then we define the visual dependency as:
\begin{equation}
\begin{aligned}
V(\mathbf{y}) &= \frac{1}{T}\sum_{t=1}^{T}
D_{\mathrm{KL}}\Big(
\pi_\theta(\cdot \mid q,I,\mathbf{y}_{<t}) \\
&\quad \| \;
\pi_\theta(\cdot \mid q,\widetilde{I},\mathbf{y}_{<t})
\Big).
\end{aligned}
\end{equation}
A higher $V(\mathbf{y})$ indicates that the trajectory is more grounded in visual evidence rather than relying solely on language priors.
Then, we consider the stability of the reasoning process and use the average trajectory entropy as its proxy:
\begin{equation}
H(\mathbf{y}) = \frac{1}{T}\sum_{t=1}^{T}
\mathcal{H}\big(
\pi_\theta(\cdot \mid q,I,\mathbf{y}_{<t})
\big),
\end{equation}
where $\mathcal{H}(p) = -\sum_x p(x)\log p(x)$ denotes the entropy of the token distribution. ExGRPO~\citep{zhan2026exgrpo} has observed that lower-entropy trajectories tend to exhibit more stable and higher-quality reasoning behaviors under RLVR.
Together, $V(\mathbf{y})$ and $H(\mathbf{y})$ provide complementary views of trajectory quality: the former captures grounding in visual evidence, while the latter reflects the stability of the reasoning process. These signals are used as complementary criteria for selecting high-quality trajectories in experience replay.

\section{Method}
\label{sec:method}

\begin{figure*}[t]
    \centering         
    \includegraphics[width=\textwidth, height=0.5\textwidth]{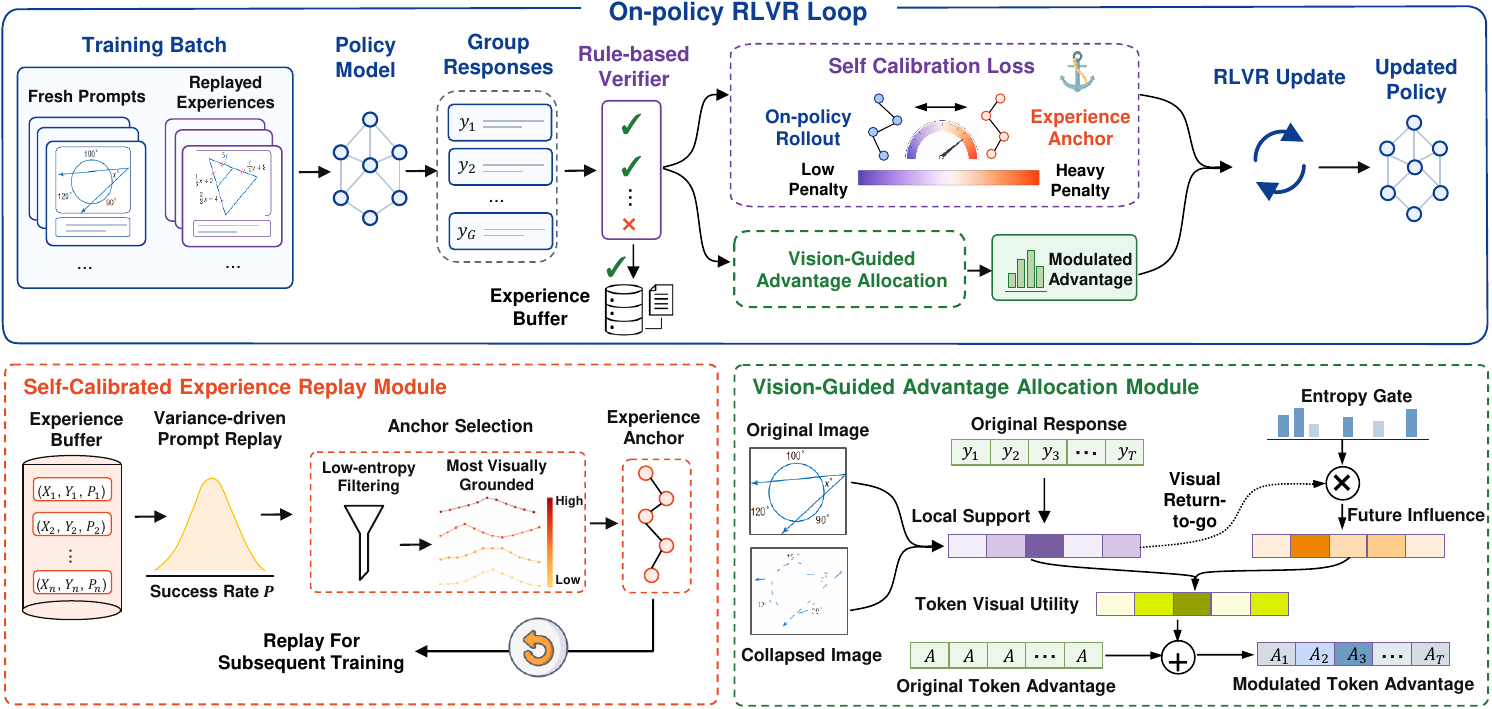}   
    \caption{Overview of the PIVOT framework. Standard RLVR is augmented by two synergistic modules: (1) \textit{Self-Calibrated Experience Replay} module selectively collects and replays valuable prompts alongside high-quality experiences to guide policy optimization through a self-calibration loss; and (2) \textit{Vision-Guided Advantage Allocation} module allocates token advantages by combining local counterfactual visual support with entropy-gated future influence to reinforce pivotal perception or reasoning steps.}
    \label{fig:overview}
\end{figure*}

In this section, we introduce PIVOT, a dual-level learning framework designed to preserve and amplify sparse yet informative learning signals for visually-grounded reasoning. Our approach extends the standard RLVR paradigm with two synergistic components: (1) a self-calibrated experience replay mechanism that collects and reuses high-quality past experiences to guide the policy optimization, and (2) a vision-guided advantage allocation strategy that reallocate token advantages to prioritize visually critical reasoning steps. These components work in tandem to transform rare successes into dense and informative signals, allowing the model to more effectively internalize visually grounded reasoning patterns. Figure~\ref{fig:overview} provides a schematic overview of our framework.

\subsection{Self-Calibrated Experience Replay}
\label{sec:trajectory_level}
This mechanism primarily addresses the issue of high-quality trajectories being discarded during on-policy sampling, which results in wasted optimization signals. It achieves this by collecting and selectively replaying high-quality trajectories and using them as reference anchors to guide policy optimization.

\paragraph{Experience Pool Curation.}
During the rollout phase, the model generates $G$ candidate responses $\{\mathbf{y}_i\}_{i=1}^G$ for a given multimodal input $x^*=(q^*, I^*)$, which are then evaluated by a rule-based verifier. We denote the subset of $k$ successful responses as $\mathcal{Y}^+=\{y_{+}^*\}$, and estimate the empirical success probability for this input as $\hat{p}(x^*) = k/G$. To efficiently manage these experiences, we store them in an experience buffer $\mathcal{B}$ as a structured mapping $x^* \mapsto \{\hat{p}(x^*), \mathcal{Y}^+\}$. This design uniquely associates each input with its empirical success probability and a set of successful responses.

\paragraph{Selective Experience Replay.}
A naive replay strategy involves randomly selecting prompts and successful trajectories from the experience pool. However, previous work~\citep{zhan2026exgrpo} has shown that this approach is suboptimal. We argue that prompts at the \textit{frontier of the model's capabilities} provide the most informative training signal. And we identify these prompts by estimating their reward variance, which has been shown to be a good proxy both theoretically and empirically~\citep{NEURIPS2025_45d58239, jiang2025vcrl}. For a multimodal prompt $x^*$ in the experience buffer, the variance can be estimated by $\hat{p}(x^*)(1-\hat{p}(x^*))$. When constructing a training batch of size $N$, we sample $N_{\text{off}}$ prompts from the experience buffer with a probability proportional to their estimated variance:
\begin{equation}
    P(\text{sample prompt } x^*) \propto \hat{p}(x^*)(1-\hat{p}(x^*))
\end{equation}
This variance-driven sampling naturally creates a curriculum that focuses the replay mechanism on partially solved problems.

For a selected prompt, the experience buffer may contain multiple successful historical trajectories. Instead of random selection, we will select a trajectory that is both visually grounded and stable under the current model. Specifically, based on the metrics defined in Section~\ref {pre:tqs}, we use the visual dependency score $V(\mathbf{y})$ to quantify how strongly a trajectory relies on visual information, and the trajectory entropy $H(\mathbf{y})$ to measure the stability of the reasoning process. We employ a two-stage filtering process to select the optimal one from the candidate set $\mathcal{Y}^+$. First, we rank all trajectories in $\mathcal{Y}^+$ by their entropy $H(\mathbf{y})$ in ascending order, and retain the top $\alpha\%$ (e.g., $\alpha = 50$) with the lowest entropy to form a stable subset $\mathcal{Y}^*$. Subsequently, among these remaining high-quality candidates, we select the one that maximizes the visual dependency score as our trajectory-level anchor $\mathbf{y}^{\text{exp}}$:
\begin{equation}
    \mathbf{y}^{\text{exp}} = \mathop{\arg\max}_{\mathbf{y} \in \mathcal{Y}^*} V(\mathbf{y})
\end{equation}

\paragraph{Experience Calibration Loss.}
After the experience sampling process described above, we have constructed a batch of $N$ prompts. Among them, $N_{\text{off}}$ prompts are selectively sampled from the experience buffer, each paired with its optimal historical response $\mathbf{y}^{\text{exp}}$. The model will perform a standard rollout phase on this batch, and each prompt will get a group of $G$ responses, denoted as $\{\mathbf{y}^{\text{cur}}_i\}_{i=1}^G$.

For prompts replayed from the experience buffer, we introduce an additional self-calibration loss to uncover valuable signals from historical experiences. Rather than treating the historical experiences as a direct optimization target, we utilize them as stable reference anchors for self-calibration. We measure the policy's relative preference between its newly generated trajectory $\mathbf{y}^{\text{cur}}_i$ and the historical experience $\mathbf{y}^{\text{exp}}$ by defining the log-confidence gap:
\begin{equation}
    \Delta \ell_i = \log \pi_\theta(\mathbf{y}^{\text{cur}}_i \mid q, I) - \text{sg} \left[ \log \pi_\theta(\mathbf{y}^{\text{exp}} \mid q, I) \right]
\end{equation}
where $\text{sg}[\cdot]$ denotes a stop-gradient operator, ensuring $\mathbf{y}^{\text{exp}}$ serves as a stable reference anchor. To dynamically guide the policy based on the correctness of its current exploration, we formulate a contrastive margin loss:
\begin{equation}
    \mathcal{L}_{\text{exp}} = |A_i| \cdot \operatorname{softplus}\left( -s \cdot \Delta \ell_i \right)
\end{equation}
where $|A_i|$ denotes the magnitude of the group-wise advantage for $\mathbf{y}^{\text{cur}}_i$, and $s$ is an indicator variable set to $+1$ if $\mathbf{y}^{\text{cur}}_i$ is correct and $-1$ if it is incorrect. In practice, we implement this objective at the token level and normalize it over all valid response tokens, providing more fine-grained calibration signals.

This objective acts as a dynamic self-calibration mechanism. For successful explorations ($s=+1$), it increases the policy's confidence in $\mathbf{y}^{\text{cur}}_i$, while the softplus operator naturally decays gradients once $\Delta \ell_i > 0$ to prevent over-optimization. Conversely, for incorrect trajectories ($s=-1$), the historical anchor acts as a dynamic threshold: the loss penalizes the flawed $\mathbf{y}^{\text{cur}}_i$ until its confidence drops safely below the proven baseline.

\subsection{Vision-Guided Advantage Allocation}
\label{sec:token_level}
Although the self-calibrated experience replay mechanism preserves and leverages high-quality historical experiences by introducing an additional calibration loss, standard RLVR still broadcasts a uniform advantage to all tokens within a trajectory, resulting in the dilution of optimization signals at the token level. We address this by modulating the token advantage based on its visual utility.

\paragraph{Quantifying Token Visual Utility.}
We begin by estimating the direct counterfactual visual support for each generated token. This is achieved by comparing the model's predictive probability under the original image \(I\) against the counterfactually corrupted image \(\widetilde{I}\) (see Section~\ref{pre:tqs}). We formulate this metric as:
\begin{equation}
    c_t =
    1 -
    \frac{
    \pi_\theta(y_t \mid q,\widetilde I,\mathbf{y}_{<t})
    }{
    \pi_\theta(y_t \mid q,I,\mathbf{y}_{<t})
    }.
\end{equation}
This formulation captures the relative change in the model's support for the sampled token after visual perturbation. A positive \(c_t\) indicates that the sampled token receives positive support from the visual evidence, suggesting it is more likely to be a perception-critical token. Conversely,  \(c_t\approx0\) or \(c_t < 0\) implies a lack of visual support, indicating that the token might be generated from language priors or visual hallucinations. Detailed justification is provided in Appendix~\ref{app:visual_support_score}.

However, using only local counterfactual support as visual utilities may overlook the impact of tokens on future visual reasoning processes. To capture this impact, we define the future-discounted visual support score:
\begin{equation}
    F_t = 
    \frac{\sum_{k > t} \gamma^{k-t-1} c_k}
    {\sum_{k > t} \gamma^{k-t-1}},
\end{equation}
where \(\gamma \in [0,1]\) is a discount factor. In implementation, the summation is truncated to a finite window of size \(W\), which removes noisy long-tail effects from distant tokens.

To avoid indiscriminately propagating future visual credit to all preceding tokens, we further introduce an entropy-based gate to highlight reasoning-pivot tokens~\citep{NEURIPS2025_a797c2d2}. Let \(H_t\) denote the token entropy and \(\bar{H}\) the average entropy over the batch. We define
\begin{equation}
    u_t = 1 - \exp\left(-\frac{H_t}{\bar{H}}\right).
\end{equation}
This gate assigns larger weights to high-entropy positions, which are more likely to be reasoning decision points. The final token visual utility is
\begin{equation}
    U_t = c_t + \lambda \cdot \operatorname{Detrending}(u_t F_t),
\end{equation}
where \(\lambda\) balances local visual support and future-aware influence. \(\operatorname{Detrending}(\cdot)\) is a standard OLS detrending operator~\citep{doi:10.1073/pnas.0701020104} to remove potential systematic position bias; implementation details are provided in Appendix~\ref{app:detrending}.

\paragraph{Advantage Modification.}
The final token visual utility score $U_t$ not only measures local visual support but also captures its impact on downstream visual reasoning. We then use it to modulate the original advantage $\hat{A}_{i,t}$ (which is uniform across tokens within the trajectory):
\begin{equation} \label{eq:adv_mod}
    \hat{A}'_{i,t} = \hat{A}_{i,t} + \beta |\hat{A}_{i,t}| U_t,
\end{equation}
where $\beta$ is a scaling factor. Then we apply the following sign protection operation to $\hat{A}' _ {i,t}$:
\begin{equation}
\hat{A} _ {i,t}^{\text{final}} =
\begin{cases}
\max(\hat{A}' _ {i,t}, 0), & \text{if } R _ i = 1, \\\\
\min(\hat{A}' _ {i,t}, 0), & \text{if } R _ i = 0.
\end{cases}
\end{equation}
This operation prevents the modulation from reversing the optimization directions of positive and negative samples. All computations within this module are detached from the gradient graph, as they are only utilized to modulate the scalar token advantages. The policy gradient is then computed using the refined token advantages.

\subsection{Final Training Objective}
Our dual-level learning framework is integrated into the standard policy optimization process. The final optimization objective combines the RLVR loss with our modulated advantages $\hat{A}^{\text{final}}_{i,t}$, and the self-calibration experience loss $\mathcal{L}_{\text{exp}}$:
\begin{equation}
    \mathcal{L}_{\text{final}}(\theta) = \mathcal{L}_{\text{RLVR}}(\theta; \hat{A}^{\text{final}}_{i,t}) + \lambda_{\text{exp}} \mathcal{L}_{\text{exp}}(\theta),
\end{equation}
where $\mathcal{L}_{\text{RLVR}}(\theta; \hat{A}^{\text{final}}_{i,t})$ is the standard RLVR loss using the modified advantage $\hat{A}^{\text{final}}_{i,t}$, and $\lambda_{\text{exp}}$ is a coefficient balancing the two components. This objective not only preserves the on-policy learning signal but also anchors policy optimization to high-quality historical experiences and visually critical tokens, improving the utilization efficiency of high-value optimization signals in RLVR.

\section{Experiments}

\subsection{Experimental Setup}
\paragraph{Models and Baselines.}
We adopt Qwen-2.5-VL-3B and Qwen-2.5-VL-7B as our base models. To evaluate data scalability, we first conducted experiments on the Geometry3K~\citep{lu-etal-2021-inter} dataset and further extended the training to the larger-scale VIRL39K dataset~\citep{NEURIPS2025_2c84844a}. We reproduce the algorithms GRPO~\citep{shao2024deepseekmathpushinglimitsmathematical}, DAPO~\citep{NEURIPS2025_a4277440}, PAPO~\citep{wang2026perceptionaware}, and VPPO~\citep{huang2026spotlight} across different model scales and datasets. 
\paragraph{Evaluation.}
The evaluation benchmarks include MathVista~\citep{lu2024mathvista}, MathVerse~\citep{10.1007/978-3-031-73242-3_10}, We-Math~\citep{qiao-etal-2025-math}, MMK12~\citep{meng2025mmeurekaexploringfrontiersmultimodal} and Geometry3K~\citep{lu-etal-2021-inter}, LogicVista~\citep{xiao2024logicvista}, SuperClevr-Counting~\citep{li2023super}, MMMU-Pro~\citep{yue-etal-2025-mmmu} and MathVerse-V~\citep{10.1007/978-3-031-73242-3_10}. We also test on ScienceQA~\citep{lu2022learn}, HallusionBench~\citep{10657594}, ChartQAPro~\citep{masry-etal-2025-chartqapro}, InfographicVQA~\citep{mathew2022infographicvqa} and RealWorldQA from lmms-eval~\citep{zhang-etal-2025-lmms} to evaluate out-of-domain performance. To reduce reliance on LLM-as-a-judge systems, we adopt an exact-match scoring protocol and report the average accuracy@8 with an inference temperature of 1.0 following PAPO~\citep{wang2026perceptionaware}. More details are provided in Appendix~\ref{app:data_and_evaluation}.
\paragraph{Implementation Details.}
Our training framework follows the DAPO~\citep{NEURIPS2025_a4277440} recipe with a learning rate of 1e-6, a rollout batch size of 384, and a maximum response length of 2048. Models are trained for 15 epochs on Geo3k and 2 epochs on VIRL39K. We set the rollout group size $n=5$ for the 3B model, and $n=8$ for the 7B models. Following VPPO~\citep{huang2026spotlight}, we apply a small entropy penalty (coefficient 0.06) to all models during training to ensure training stability and fair comparison. More details are provided in Appendix~\ref{app:Implementation_details}.

\begin{table*}[ht]
\centering
\setlength{\tabcolsep}{2.5pt}
\renewcommand{\arraystretch}{1.2}
\definecolor{gainred}{RGB}{230,60,60}
\newcommand{\gain}[1]{\textcolor{gainred}{(+#1)}}
{\small
\begin{tabular}{l|ccccc|ccc|c}
\toprule
\multicolumn{1}{c|}{\multirow{2}{*}{\textbf{Model}}} & \multicolumn{5}{c|}{\textbf{Mathematical \& Geometric Reasoning}} & \multicolumn{3}{c|}{\textbf{Vision-Dependent Reasoning}} & \multirow{2}{*}{\textbf{Avg.}} \\
\cmidrule(lr){2-6} \cmidrule(lr){7-9}
 & \textbf{Geo3K\textsubscript{test}} & \textbf{MathVerse} & \textbf{MathVista} & \textbf{WeMath} & \textbf{MMK12} & \textbf{LogicVista} & \textbf{Counting} & \textbf{MathVerse\textsubscript{$V$}} &  \\
\midrule
\textit{Qwen2.5-VL-3B}           & 18.95 & 33.33 & 46.49 & 30.10 & 36.47 & 29.42 & 37.25 & 30.55 & 32.82 \\
\quad + GRPO                     & 35.88 & 42.92 & 52.43 & 47.25 & 40.57 & 36.66 & 47.25 & 39.28 & 42.78 \\
\quad + DAPO                     & 41.60 & 46.88 & 55.19 & 53.00 & 40.53 & 37.11 & 50.75 & 43.38 & 46.06 \\
\quad + PAPO\textsubscript{\textit{D}}                     & \textbf{43.68} & 50.46 & \underline{55.54} & 54.87 & 40.92 & \underline{38.17} & \underline{56.31} & \underline{47.59} & 48.44 \\
\quad + VPPO                     & \underline{43.16} & \underline{51.89} & 54.28 & \underline{58.05} & \underline{42.14} & 37.78 & 55.31 & 46.78 & \underline{48.67} \\
\rowcolor{gray!12}
\quad \textbf{+ PIVOT (Ours)}             & 43.07 & \textbf{53.90} & \textbf{56.25} & \textbf{60.07} & \textbf{43.05} & \textbf{38.20} & \textbf{65.63} & \textbf{49.83} & \textbf{51.25} \\
\midrule
\textit{Qwen2.5-VL-7B}           & 35.73 & 39.02 & 63.50 & 45.80 & 43.11 & 42.84 & 66.75 & 34.58 & 46.42 \\
\quad + GRPO                     & 47.40 & 47.84 & 66.46 & \underline{56.61} & 45.52 & 43.82 & 72.69 & 43.56 & 52.99 \\
\quad + DAPO                     & 51.73 & 43.77 & 65.56 & 49.73 & 43.74 & \underline{46.09} & 80.50 & 37.41 & 52.32 \\
\quad + PAPO\textsubscript{\textit{D}}                     & 51.66 & 48.70 & 65.96 & 54.27 & 44.73 & 45.69 & \textbf{84.56} & 43.14 & 54.84 \\
\quad + VPPO                     & \underline{52.06} & \underline{51.03} & \textbf{68.94} & 55.78 & \underline{47.11} & 44.97 & 80.13 & \underline{44.24} & \underline{55.53} \\
\rowcolor{gray!12}
\quad \textbf{+ PIVOT (Ours)}             & \textbf{53.14} & \textbf{54.44} & \underline{68.38} & \textbf{60.88} & \textbf{49.64} & \textbf{46.28} & \underline{82.50} & \textbf{47.52} & \textbf{57.85} \\
\bottomrule
\end{tabular}
}
\caption{Performance comparison of PIVOT against various baselines. Models are trained on the Geometry3K dataset using Qwen2.5-VL-3B and 7B as backbones. \textbf{Bold} and \underline{underlined} indicate the best and second-best results.}
\label{tab:geo3k_results}
\end{table*}

\begin{table*}[ht]
\centering
\setlength{\tabcolsep}{2.5pt}
\renewcommand{\arraystretch}{1.2}
\definecolor{gainred}{RGB}{230,60,60}
\newcommand{\gain}[1]{\textcolor{gainred}{(+#1)}}
{\small
\begin{tabular}{l ccccccccc c}
\toprule
\textbf{Method} & \textbf{Geo3K} & \textbf{MathVista} & \textbf{MathVerse} & \textbf{WeMath} & \textbf{MMK12} & \textbf{LogicVista} & \textbf{Counting} & \textbf{MMMU-Pro} & \textbf{MathVerse\textsubscript{$V$}} & \textbf{Avg.} \\
\midrule
Base                & 19.22 & 46.49 & 33.33 & 30.10 & 36.47 & 29.42 & 37.25 & 19.66 & 30.55 & 31.39 \\
\midrule
GRPO                & 30.99 & 59.15 & 55.37 & 60.05 & 58.18 & 39.71 & 58.81 & 27.06 & 52.08 & 49.04 \\
DAPO                & 36.29 & 60.14 & 60.87 & 62.97 & 62.11 & 42.51 & 75.44 & 28.32 & 57.23 & 53.99 \\
PAPO\textsubscript{\textit{D}}                & \underline{37.44} & \underline{61.45} & \underline{61.88} & \textbf{65.48} & \textbf{62.57} & \underline{43.85} & \underline{77.13} & \underline{28.36} & \underline{58.92} & \underline{55.23} \\
VPPO                & 36.04 & 60.33 & 59.78 & 64.18 & \underline{62.26} & 42.25 & 74.56 & 28.14 & 56.95 & 53.83 \\
\rowcolor{gray!12}
\textbf{PIVOT} & \textbf{37.73} & \textbf{63.04} & \textbf{62.52} & \underline{65.01} & 61.96 & \textbf{44.44} & \textbf{77.81} & \textbf{28.98} & \textbf{59.67} & \textbf{55.68} \\
\bottomrule
\end{tabular}
}
\caption{Performance evaluation of PIVOT and baselines trained on the larger-scale VIRL39K dataset using Qwen2.5-VL-3B as base model. \textbf{Bold} and \underline{underlined} indicate the best and second-best results.}
\label{tab:virl39k_results}
\end{table*}

\subsection{Main Results}
\paragraph{Multimodal Reasoning Performance.} 
Table~\ref{tab:geo3k_results} and Table~\ref{tab:virl39k_results} present the performance of PIVOT compared to other baselines. When trained on the Geometry3K dataset, PIVOT consistently achieves the highest average accuracy for both the 3B and 7B models, reaching 51.25 (11.27\% relative gains over DAPO) and 57.85 (10.57\% relative gains over DAPO), respectively. The performance improvements are observed in both general mathematical and vision-dependent reasoning tasks. To validate data scalability, we further extend the training of the 3B model to the larger-scale VIRL39K dataset. As shown in Table~\ref{tab:virl39k_results}, PIVOT maintains its superiority, yielding the highest average score of 55.68 and outperforming competitive baselines across most evaluated benchmarks. These results demonstrate the effectiveness of PIVOT in enhancing the multimodal reasoning capabilities of LVLMs.

\begin{figure}[htp]
  \centering
  \begin{subfigure}{0.49\columnwidth}
    \centering
    \includegraphics[width=\linewidth]{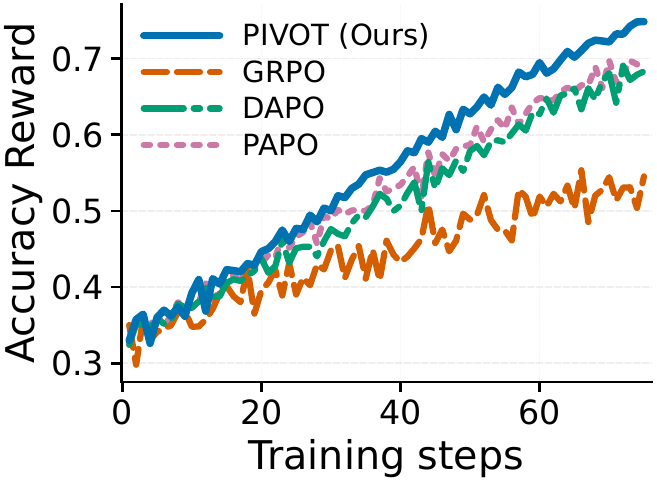}
    \caption{Accuracy Rewards.}
    \label{fig:left}
  \end{subfigure}
  \hfill
  \begin{subfigure}{0.49\columnwidth}
    \centering
    \includegraphics[width=\linewidth]{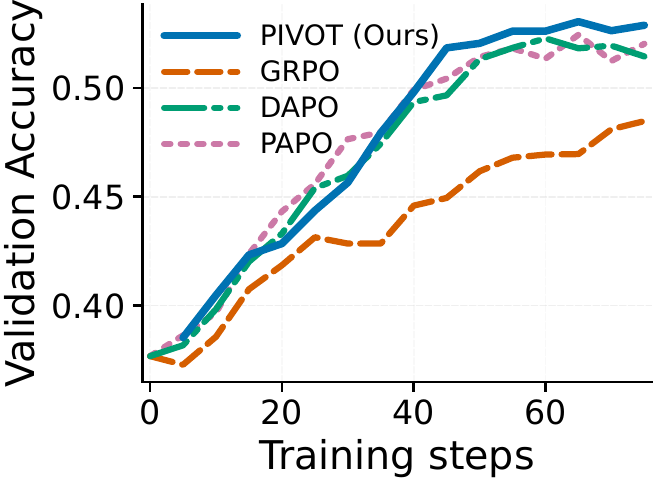}
    \caption{Validation Accuracy.}
    \label{fig:right}
  \end{subfigure}
  \caption{Training dynamics of Qwen2.5-VL-7B on the Geometry3k dataset: (a) training accuracy rewards, and (b) validation accuracy on the Geometry3k@test data~\citep{lu-etal-2021-inter}.}
  \label{fig:training_dynamics}
\end{figure}

\paragraph{Training Dynamics.} 
Figure~\ref{fig:training_dynamics} illustrates the training dynamics of the 7B model on Geo3k dataset. As shown in Figure~\ref{fig:training_dynamics}\subref{fig:left}, PIVOT demonstrates superior learning efficiency, achieving higher training accuracy rewards compared to other baselines. Furthermore, the corresponding validation accuracy (Figure~\ref{fig:training_dynamics}\subref{fig:right}) confirms that this efficient optimization directly translates into better generalization results on unseen validation data. These improvements demonstrate that our proposed method provides a more robust and effective optimization process. 

\subsection{Quantitative Analysis}

\paragraph{Ablation Study.}

\begin{table*}[htp]
\centering
\setlength{\tabcolsep}{1.4pt}
\renewcommand{\arraystretch}{1.2}
\definecolor{gainred}{RGB}{230,60,60}
\definecolor{declinegreen}{RGB}{60,130,60}
\newcommand{\gain}[1]{\textcolor{gainred}{+#1}}
\newcommand{\decline}[1]{\textcolor{declinegreen}{-#1}}
{\small
\begin{tabular}{l cccccccc cc}
\toprule
\textbf{Variant} & \textbf{Geo3k\textsubscript{test}} & \textbf{MathVerse} & \textbf{MathVista} & \textbf{WeMath} & \textbf{MMK12} & \textbf{LogicVista} & \textbf{Counting} & \textbf{MathVerse\textsubscript{$V$}} & \textbf{Avg.} & \textbf{$\Delta$} \\
\midrule
DAPO (Baseline)      & 41.60 & 46.88 & 55.19 & 53.00 & 40.53 & 37.11 & 50.75 & 43.38 & 46.06 & -- \\
\quad + SER only     & 42.37 & 51.19 & 55.23 & 57.67 & 42.51 & 38.62 & 55.38 & 48.05 & 48.88 & \gain{6.12\%} \\            
\quad + VAA only     & 44.09 & 48.56 & 57.08 & 57.51 & 41.74 & 38.93 & 65.94 & 45.39 & 49.91 & \gain{8.36\%} \\
\midrule
Off-PG + VAA         & 44.20 & 51.06 & 56.89 & 54.31 & 42.18 & 39.23 & 67.75 & 46.99 & 50.33 & \gain{9.17\%} \\
Off-SFT + VAA        & 44.03 & 51.26 & 55.88 & 55.94 & 41.91 & 38.95 & 61.50 & 46.78 & 49.53 & \gain{7.10\%} \\
SER + VAA\textsubscript{local}     & 44.57 & 49.76 & 55.68 & 57.52 & 42.43 & 38.42 & 59.38 & 45.60 & 49.17 & \gain{6.75\%} \\
CalibRL              & 41.43 & 44.75 & 52.85 & 50.79 & 38.70 & 36.52 & 56.81 & 40.75 & 45.33 & \decline{1.58\%} \\
CalibRL+VAA          & 44.05 & 49.59 & 54.86 & 54.56 & 41.06 & 37.98 & 60.00 & 45.75 & 48.48 & \gain{5.25\%} \\
\midrule
\textbf{PIVOT (full)}      & 43.07 & 53.90 & 56.25 & 60.07 & 43.05 & 38.20 & 65.63 & 49.83 & \textbf{51.25} & \gain{11.27\%} \\
\bottomrule
\end{tabular}
}
\caption{Ablation of Self-calibrated Experience Replay (SER) module and Vision-guided Advantage Allocation (VAA) module. Their combination yields the best results, confirming the effectiveness of our dual-level framework.}
\label{tab:module_ablation}
\end{table*}

To isolate the contributions of different modules in our framework, we conduct ablation studies using the Qwen2.5-VL-3B model trained on the Geo3k dataset. PIVOT consists of two core modules: Self-calibrated Experience Replay (SER) and Vision-guided Advantage Allocation (VAA). We ablate each module to verify its influence. Moreover, we further conduct ablations on the replay and token-utility designs. Specifically, \textit{Off-PG+VAA} reuses successful experiences via off policy gradient updates, whereas \textit{Off-SFT+VAA} uses them as supervised fine-tuning targets. 
In addition, \textit{SER+VAA\textsubscript{local}} removes the entropy-gated future visual utility from VAA, isolating the effect of local visual utility alone. As shown in Table~\ref{tab:module_ablation}, both SER-only and VAA-only yields noticeable performance gains over the baseline, and their combination yields the best average performance across evaluation benchmarks. \textit{Off-PG+VAA} and \textit{Off-SFT+VAA} improve over baseline but lag behind full PIVOT, indicating the advantage of using high-quality experiences as calibration anchors. The gap between \textit{SER+VAA\textsubscript{local}} and full PIVOT further validates the contribution of entropy-gated future visual utility. It is worth noting that PIVOT slightly sacrifices the in-domain Geo3K test set performance relative to several ablations while improving broader generalization. This arises from SER’s regularization against source-domain over-specialization and VAA’s modeling of downstream visual impact beyond local token–image correspondence. Together, these designs balance exploitation and exploration, enabling broader generalization.

In addition, to isolate the specific benefits of our self-generated visually grounded anchors, we also conducted experiments evaluating both CalibRL and CalibRL+VAA. As shown in Table~\ref{tab:module_ablation}, PIVOT consistently outperforms CalibRL and the CalibRL+VAA variant on average and across almost all benchmarks. This superiority can be attributed to two main factors: (1) PIVOT's self-generated anchors, which are inherently compatible with the evolving policy to ensure better-calibrated confidence comparisons, and (2) the proposed entropy and visual-dependency filtering mechanisms. We also ablate different visual intervention strategies; details are shown in Appendix~\ref{app:ablation_intervention}.

\paragraph{Out-of-Domain Generalization.}
To ensure that our model is not simply over-optimized for math and geometry benchmarks, we evaluate its Out-of-Domain generalization capabilities on ScienceQA, HallusionBench, ChartQAPro, InfographicVQA, and RealWorldQA. As shown in Table~\ref{tab:ood}, our method consistently achieves improvements over baselines across different OOD datasets. These results suggest that the visual reasoning capabilities acquired through our training paradigm can generalize to broader multimodal tasks and help mitigate multimodal hallucinations.

\begin{table*}[t]
\centering
{%
\footnotesize
\setlength{\tabcolsep}{2pt}
\renewcommand{\arraystretch}{1.18}
\begin{tabular}{l c c c c c c}
\toprule
\textbf{Method} & \textbf{ScienceQA} & \textbf{HallusionBench} & \textbf{ChartQAPro} & \textbf{InfographicVQA} & \textbf{RealWorldQA} & \textbf{Avg.} \\
\midrule
Qwen2.5-VL-3B-Instruct  & 68.75 & 51.76 & 26.90 & 49.73 & 44.59 & 48.35 \\
\quad + DAPO            & 80.25 & 56.02 & 32.94 & 51.64 & 46.90 & 53.55 \\
\quad \textbf{+ PIVOT (Ours)} & \textbf{82.04} & \textbf{57.64} & \textbf{34.73} & \textbf{52.86} & \textbf{47.47} & \textbf{54.95} \\
\bottomrule
\end{tabular}
}
\caption{Out-of-Domain generalization results. Though trained only on the Geometry3K dataset, our method shows strong generalization to out-of-domain multimodal tasks.}
\label{tab:ood}
\end{table*}

\begin{table*}[htp]
\centering
\setlength{\tabcolsep}{1.7pt}
\renewcommand{\arraystretch}{1.2}
\definecolor{gainred}{RGB}{230,60,60}
\newcommand{\gain}[1]{\textcolor{gainred}{+#1}}
{\small
\begin{tabular}{l cccccccc cc}
\toprule
\textbf{Strategy} & \textbf{Geo3K} & \textbf{MathVerse} & \textbf{MathVista} & \textbf{WeMath} & \textbf{MMK12} & \textbf{LogicVista} & \textbf{Counting} & \textbf{MathVerse\textsubscript{$V$}} & \textbf{Avg} & $\Delta$ \\
\midrule
Qwen2.5-VL-3B & 18.95 & 33.33 & 46.49 & 30.10 & 36.47 & 29.42 & 37.25 & 30.55 & 32.82 & -- \\
\midrule
GRPO (Baseline)      & 35.88 & 42.92 & 52.43 & 47.25 & 40.57 & 36.66 & 47.25 & 39.28 & 42.78 & \gain{30.35\%} \\
\textbf{PIVOT}\textsubscript{G}  & \textbf{37.79} & \textbf{45.29} & \textbf{55.04} & \textbf{47.57} & \textbf{41.81} & \textbf{37.00} & \textbf{53.94} & \textbf{42.84} & \textbf{45.16} & \gain{37.60\%} \\
\bottomrule
\end{tabular}
}
\caption{Results of extending PIVOT to the GRPO algorithm.}
\label{app:grpo_pivot}
\end{table*}

\begin{figure}[t]
  \includegraphics[width=\columnwidth]{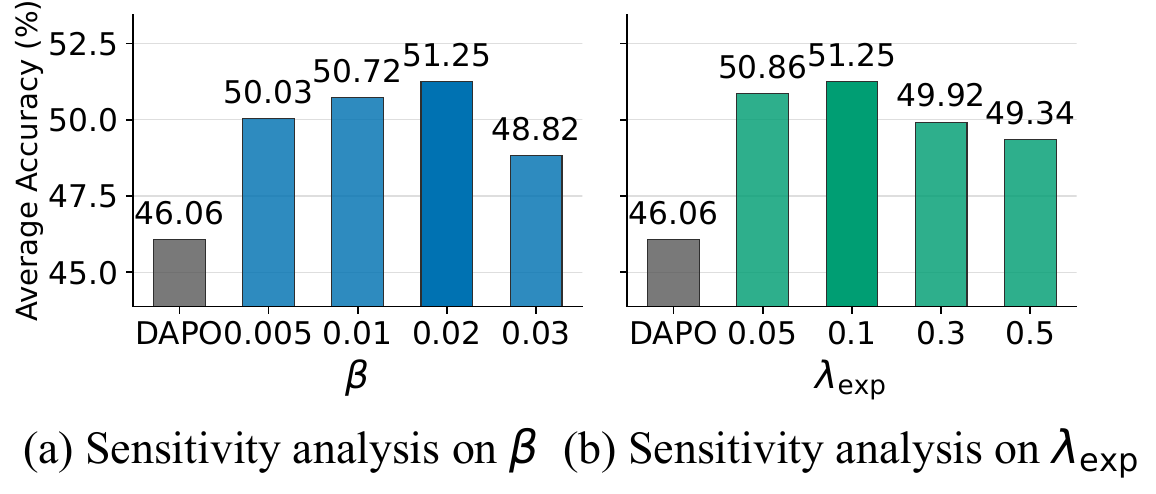}
  \caption{Sensitivity Analysis of $\beta$ and $\lambda_{\text{exp}}$.}
  \label{fig:sensitivity_analysis}
\end{figure}

\paragraph{Sensitivity Analysis}
To evaluate the robustness of our method, we conduct sensitivity analyses on two key hyperparameters: the advantage scaling factor $\beta$ and the loss balancing coefficient $\lambda_{\text{exp}}$. As shown in Figure~\ref{fig:sensitivity_analysis}, both hyperparameters exhibit a similar pattern: moderate values lead to the best performance. This is consistent with our design intuition. A small $\beta$ weakens the effect of visual advantage modulation, whereas a large $\beta$ may over-amplify visual signals and destabilize optimization. Similarly, $\lambda_{\text{exp}}$ needs to balance the experience calibration loss with the main RL objective: insufficient weighting limits its benefit, while excessive weighting may interfere with on-policy learning.

\paragraph{Generalize to GRPO}
To verify the generalizability of PIVOT, we extended our training recipe to Group Relative Policy Optimization (GRPO) \citep{shao2024deepseekmathpushinglimitsmathematical}, which serves as another widely adopted baseline in the RLVR paradigm. Specifically, we seamlessly integrated our self-calibrated experience replay and vision-guided advantage modulation into the standard GRPO optimization loop. 

As shown in Table~\ref{app:grpo_pivot}, incorporating PIVOT consistently improves standard GRPO's performance across multimodal reasoning tasks. This consistent improvement indicates that our dual-level learning framework is highly orthogonal to the choice of the specific RLVR algorithms. Rather than being confined to a specific algorithm, PIVOT functions as a general framework that effectively enhances the visual grounding capability of LVLMs during reinforcement learning.


\subsection{Additional Results}
We also provide several additional results to complement the main experiments. Specifically, Appendix~\ref{app:token_visualization} provides the token visual utility visualizations, which demonstrate that PIVOT concentrates the RL optimization signal on visually grounded perception tokens and reasoning pivots that are more relevant to solving the problem. Appendix~\ref{app:attention_analysis} presents a layer-wise attention analysis, which provides complementary evidence that PIVOT encourages the model to attend more to visual tokens when performing multimodal reasoning. Appendix~\ref{app:overhead} reports the computational overhead of the proposed framework. These results offer further insights into the behavior and practical cost of PIVOT.

\section{Conclusion}

In this paper, we proposed PIVOT, a dual-level learning framework that addresses key optimization bottlenecks in multimodal RLVR by preserving and amplifying visual reasoning signals. The framework collects and selectively replays high-quality experiences as optimization anchors through a self-calibration loss. Building upon this, PIVOT further allocates fine-grained visual advantages to tokens based on their local visual support and downstream impact on reasoning. Extensive experiments demonstrate that PIVOT consistently improves multimodal reasoning capabilities and out-of-domain generalization of LVLMs. Future work could consider extending this dual-level learning framework to broader multimodal agent tasks.

\section*{Limitations}

Despite the promising performance of PIVOT, several limitations remain to be addressed in future work. First, the self-calibrated experience replay module and the counterfactual intervention strategy inevitably introduce additional computational and storage overhead during training. Although these overheads are acceptable, exploring more resource-efficient designs and lightweight perturbation strategies remains an important direction for future optimization. Secondly, due to constrained computational resources, we primarily evaluated our framework on vision-language models up to the 7B parameter scale. While PIVOT demonstrates consistent improvements on these models, validating our framework on much larger scales, such as 32B variants, would be beneficial to fully verify its scalability. Finally, our current selection of high-quality trajectories relies on empirical proxies, specifically visual dependency and trajectory entropy. While these metrics are effective in practice, they may not fully capture the semantic correctness of complex multi-step reasoning. Future work could integrate structured, rubric-based evaluation criteria~\citep{gunjal2026rubrics} to define and filter high-quality visual reasoning paths more robustly.

\section*{Ethical Considerations}

This work focuses on improving multimodal reasoning capabilities of large vision-language models. All experiments are conducted on public datasets, without using private user data, personally identifiable information, or human-subject data. Nevertheless, models trained with our method may still produce incorrect or biased outputs, especially under distribution shifts or in high-stakes real-world scenarios. Thus, such models should be carefully validated and used with human oversight before deployment in domains such as medicine, law, or safety-critical inspection. AI-assisted tools were used only for language polishing and coding assistance. 

\section*{Acknowledgments}

This work is jointly supported by the National Key R\&D Program of China (Grant No. 2024YFF0505703) and Beijing Municipal Natural Science Foundation (Grant No. L257009).

\bibliography{main}

\appendix
\newpage
\section{Justification of the Counterfactual Visual Support Score}
\label{app:visual_support_score}

In Section~\ref{sec:token_level}, we define the direct counterfactual support score as a probability ratio rather than a standard log-likelihood difference. Here, we provide the theoretical justification for this design choice. 

For simplicity of notation at generation step $t$, let $p_t(y) = \pi_\theta(y_t=y \mid q, I, \mathbf{y}_{<t})$ denote the token distribution conditioned on the original image, and $q_t(y) = \pi_\theta(y_t=y \mid q, \widetilde{I}, \mathbf{y}_{<t})$ denote the distribution conditioned on the corrupted image. For a sampled token $y_{i,t}$, our metric is defined as:
\begin{equation}
    c_t = 1 - \frac{q_t(y_{t})}{p_t(y_{t})}.
\end{equation}
And this computation is detached from the gradient graph. Intuitively, this quantity measures whether the sampled token receives more probability support from the original-image condition than from the corrupted-image condition. If \(p_t(y_{t})>q_t(y_{t})\), then \(c_{t}>0\), meaning that the token is more likely to be generated when the correct visual evidence is available. Such a token can be regarded as visually supported. If \(p_t(y_{t})\approx q_t(y_{t})\), then \(c_{t}\approx 0\), indicating that the token can be similarly explained with or without the original image, and is therefore more likely to correspond to language priors, formatting tokens, or generic connectors. If \(p_t(y_{t})<q_t(y_{t})\), then \(c_{t}<0\), suggesting that the token is even more favored under the corrupted-image condition, these tokens may correspond to visual hallucinations or some completely unrelated tokens and should be suppressed.

The key advantage of this form is that it estimates a signed distributional difference between the original-image and corrupted-image policies. Specifically, for any vocabulary token \(a\), we have
\begin{equation}
\begin{aligned}
    \mathbb{E}_{y\sim p_t}
    \left[
        c_t(y)\mathbf{1}\{y=a\}
    \right]
    &=
    p_t(a)
    \left(
        1-\frac{q_t(a)}{p_t(a)}
    \right)  \\
    &=
    p_t(a)-q_t(a).
\end{aligned}
\end{equation}
Thus, under sampling from the original-image policy, the residual provides an unbiased estimate of the signed probability-mass difference \(p_t-q_t\). 

This property also explains why the score is preferable to directly using a log-likelihood gap such as
\begin{equation}
    \log p_t(y_{t})-\log q_t(y_{t}).
\end{equation}
Although the log-likelihood gap is intuitive, its expectation under \(p_t\) is always non-negative:
\begin{equation}
    \mathbb{E}_{y\sim p_t}
    \left[
        \log \frac{p_t(y)}{q_t(y)}
    \right]
    =
    D_{\mathrm{KL}}(p_t\Vert q_t)
    \geq 0.
\end{equation}
Therefore, if it is directly added to the token-level advantage, it tends to introduce a positive bonus on average, which may encourage longer responses or over-reward tokens from prefixes with large distributional divergence. By contrast, the proposed residual satisfies
\begin{equation}
\begin{aligned}
    \mathbb{E}_{y\sim p_t}
    \left[
        1-\frac{q_t(y)}{p_t(y)}
    \right]
    &=
    \sum_y p_t(y)
    -
    \sum_y q_t(y)  \\
    &=
    0 .
\end{aligned}
\end{equation}
Hence, \(c_t\) is a zero-mean signed residual under the original-image distribution. It redistributes optimization strength from tokens more favored by the corrupted-image or language-prior condition to tokens more favored by the original-image condition, rather than injecting an additional uniformly positive reward.

When used as an additive token-level advantage correction,
\begin{equation}
    \widetilde A_{i,t}
    =
    A_i + \eta r_{c,t},
\end{equation}
the score induces a policy-gradient correction that pushes the model toward the excess probability mass of the original-image distribution. To see this, consider the expected correction term for the logit of a vocabulary token \(a\). Since
\begin{equation}
    \mathbb{E}_{y\sim p_t}[c_t(y)] = 0,
\end{equation}
we have
\begin{equation}
\begin{aligned}
    &\mathbb{E}_{y\sim p_t}
    \left[
        c_t(y)\nabla_{z_t(a)}\log p_t(y)
    \right] \\
    &=
    \mathbb{E}_{y\sim p_t}
    \left[
        c_t(y)
        \big(
            \mathbf{1}\{y=a\}-p_t(a)
        \big)
    \right] \\
    &=
    \mathbb{E}_{y\sim p_t}
    \left[
        c_t(y)\mathbf{1}\{y=a\}
    \right]
    -
    p_t(a)
    \mathbb{E}_{y\sim p_t}[c_t(y)] \\
    &=
    p_t(a)-q_t(a).
\end{aligned}
\end{equation}
Therefore, this correction increases the logits of tokens whose probability mass is higher under the original-image condition and decreases those whose probability mass is higher under the corrupted-image condition. This gives \(c_{i,t}\) a clear optimization meaning: it encourages the model to reinforce the probability mass specifically contributed by visual evidence, while suppressing probability mass supported by language priors or visual hallucinations.

\section{Details of the Detrending Operator}
\label{app:detrending}

During preliminary experiments, we find that the future-aware visual score may introduce a mild length-related bias. Although the future score is computed within a finite window, tokens near different relative positions can still have systematically different future-support statistics due to sequence boundaries, response length variation, and heterogeneous reasoning structures. In some cases, this bias encourages overly long responses. To mitigate this issue, we apply a lightweight detrending step to the entropy-gated future visual score.
Let $a_{i,t}=u_{i,t}F_{i,t}$ denote the raw entropy-gated future visual score for token \(t\) in response \(i\). Let \(E_i\) be the set of valid tokens in response \(i\) and \(n_i=|E_i|\). For each token \(t\in E_i\), we define its relative position as
\begin{equation}
r_{i,t}
=
\frac{\operatorname{rank}_{E_i}(t)}{\max(n_i-1,1)},
\end{equation}
where \(\operatorname{rank}_{E_i}(t)\in\{0,\ldots,n_i-1\}\) is the rank of token \(t\) among valid tokens. Thus, \(r_{i,t}\in[0,1]\) normalizes token position within each response.

For each response, we fit a response-specific linear trend between the raw future score and the relative position:
\begin{equation}
(\alpha_i,\beta_i)
=
\arg\min_{\alpha,\beta}
\sum_{t\in E_i}
\left(
a_{i,t}-\alpha-\beta r_{i,t}
\right)^2.    
\end{equation}
The detrending operator is then defined as:
\begin{equation}
\operatorname{Detrending}(a_{i,t})
=
a_{i,t}
-
(\alpha_i+\beta_i r_{i,t}).    
\end{equation}

This operation can be viewed as response-wise position detrending. It removes the component of the future-aware score that can be explained by relative token position, and preserves the token-specific deviation beyond this trend. Therefore, the retained signal does not simply reward a token for appearing in a favorable position of the response, but reflects whether the token has stronger future visual support than expected from its location.

\section{Experimental Settings}
\label{app:exp_setting}

\subsection{Data and Evaluation}
\label{app:data_and_evaluation}

\paragraph{Training Datasets.}
To examine data scalability, we conduct reinforcement learning on two training sets with different scales and data coverage. 
\begin{itemize}
    \item Geometry3K~\citep{lu-etal-2021-inter}: a geometry problem-solving dataset containing 3,002 multiple-choice geometry problems. The dataset requires models to jointly parse geometric diagrams, understand problem statements, and perform symbolic geometric reasoning. We follow the standard split, using the 2,101 training problems for training and the 601 test problems for validation.
    \item ViRL39K~\citep{NEURIPS2025_2c84844a}: a larger-scale vision-language reinforcement learning dataset consisting of 38,870 verifiable question-answering instances. ViRL39K is constructed from newly collected problems and existing multimodal reasoning datasets through cleaning, reformatting, rephrasing, and rule-based verification. Compared with Geometry3K, it covers a broader range of visual reasoning scenarios, including grade-school problems, STEM and social topics, charts, diagrams, tables, documents, and spatial reasoning.
\end{itemize}

\paragraph{Evaluation Benchmarks.}
We conduct evaluation on diverse benchmarks to evaluate multimodal reasoning performance, details are as follows:
\begin{itemize}
    \item MathVista~\citep{lu2024mathvista}: a visual mathematical reasoning benchmark with 6,141 examples collected from 28 existing multimodal datasets and three newly created datasets. It evaluates mathematical reasoning in diverse visual contexts and requires fine-grained visual understanding as well as compositional reasoning. 
    \item MathVerse~\citep{10.1007/978-3-031-73242-3_10}: it contains 2,612 visual math problems, each reformulated into multiple modality variants with different amounts of textual and visual information. This design allows the benchmark to diagnose whether a model truly uses diagrams rather than relying only on textual cues. 
    \item We-Math~\citep{qiao-etal-2025-math}: it consists of 6.5K visual math problems, spanning 67 hierarchical knowledge concepts and five levels of knowledge granularity. It is designed to evaluate not only final-answer accuracy but also the underlying knowledge acquisition and generalization behavior of multimodal models.
    \item MMK12~\citep{meng2025mmeurekaexploringfrontiersmultimodal}: it is a K12 multimodal reasoning dataset that covers mathematics, physics, chemistry, and biology. Its evaluation set contains 2,000 multiple-choice questions, with 500 questions for each subject, and is designed to assess multidisciplinary visual reasoning under human-verified solutions. 
    \item LogicVista~\citep{xiao2024logicvista}: it is a visual logical reasoning benchmark containing 448 human-annotated multiple-choice questions. 
    \item SuperCLEVR-Counting~\citep{li2023super}: it is derived from Super-CLEVR, a controllable visual reasoning benchmark designed to diagnose domain robustness under factors such as visual complexity, question redundancy and concept compositionality. We use its counting subset to evaluate fine-grained object perception and numerical reasoning. 
    \item MMMU-Pro~\citep{yue-etal-2025-mmmu}: it is a more robust version of MMMU that filters out questions answerable by text-only models, augments answer choices from four to ten, and introduces a vision-only setting where questions are embedded into screenshots or photos.
    \item MathVerse-V~\citep{10.1007/978-3-031-73242-3_10}: it is a vision-centric subset of MathVerse, where solving the problem requires substantial information from the visual input.
\end{itemize}

\subsection{Implementation Details.}
\label{app:Implementation_details}

We trained Qwen2.5-VL-3B and Qwen2.5-VL-7B models on the Geometry3K and VIRL39K datasets. All experiments were conducted utilizing PyTorch 2.6.0 and CUDA 12.4. Following GDPO~\citep{liu2026gdpo}, we add a small group decoupled format advantage with an advantage weight of 0.1 to regularize the format of model responses. To perform intervention on images, we perform random patch masking (patch size 14, probability 0.6) on the input images following PAPO~\citep{wang2026perceptionaware}. For future visual score computation, we use a discounted future window size $W=32$, a discount factor $\gamma=0.8$, and a future coefficient $\lambda=0.5$. For training stability, the self-calibrated experience replay module is activated once the task solved ratio exceeds 0.45 or after a maximum number of warmup steps. Upon activation, 50\% of each training batch is sampled from the experience buffer. We retain the lowest-entropy half of the candidates per prompt (retaining at least one), and then select the final experience that maximizes the trajectory dependency. For all training and evaluation experiments, we used the single, standardized prompt template shown below.

\begin{tcolorbox}[
    colback=promptbg,         
    colframe=black,           
    boxrule=0.8pt,            
    arc=4pt,                  
    left=10pt, right=10pt,    
    top=8pt, bottom=8pt,      
    title=\textbf{Reasoning Template},
    coltitle=black,
    halign title=center,
    colbacktitle=promptbg,    
    titlerule=0pt,            
    fonttitle=\bfseries
]

\textbf{SYSTEM:}\\
You are a helpful assistant.

\vspace{6pt}

\textbf{USER:}\\
\textcolor{red}{\{question\}}

\vspace{6pt}

You first think through the reasoning process as an internal monologue, enclosed within <think> </think> tags. Then, provide your final answer enclosed within \textbackslash boxed\{\ \}.

\end{tcolorbox}

\section{Additional Results}

\subsection{Ablation on Intervention Strategy}
\label{app:ablation_intervention}

\begin{table*}[htp]
\centering
\setlength{\tabcolsep}{1.7pt}
\renewcommand{\arraystretch}{1.2}
\definecolor{gainred}{RGB}{230,60,60}
\newcommand{\gain}[1]{\textcolor{gainred}{+#1}}
{\small
\begin{tabular}{l cccccccc c}
\toprule
\textbf{Strategy} & \textbf{Geo3K} & \textbf{MathVerse} & \textbf{MathVista} & \textbf{WeMath} & \textbf{MMK12} & \textbf{LogicVista} & \textbf{Counting} & \textbf{MathVerse\textsubscript{$V$}} & \textbf{Avg}  \\
\midrule
DAPO (Baseline)      & 41.60 & 46.88 & 55.19 & 53.00 & 40.53 & 37.11 & 50.75 & 43.38 & 46.06 \\
\midrule
Complete Grey Mask   & 44.24 & 47.67 & 57.63 & 55.85 & 42.75 & 39.21 & 65.69 & 43.86 & 49.61 \\            
Gaussian Noise       & 44.30 & 52.87 & 56.06 & 60.02 & 43.17 & 40.24 & 64.56 & 48.69 & 51.24 \\
\midrule
\textbf{Random Patch Masking}\textsuperscript{$\dagger$}     & 43.07 & 53.90 & 56.25 & 60.07 & 43.05 & 38.20 & 65.63 & 49.83 & \textbf{51.25} \\
\bottomrule
\end{tabular}
}
\caption{Comparison of three visual intervention strategies. $^{\dagger}$ Our final choice.}
\label{app:visual_intervention}
\end{table*}


We compare three visual intervention strategies for computing visual support scores, including complete grey masking, additive Gaussian noise, and random patch masking following VPPO~\citep{huang2026spotlight}. 
\begin{itemize}
    \item Complete masking replaces the entire image with a neutral grey canvas with RGB value $(128,128,128)$, removing all visual information. 
    \item Gaussian noise adds pixel-wise noise with a standard deviation of $189$, which is calibrated such that each pixel has approximately a $50\%$ probability of being saturated to its maximum or minimum value. 
    \item Random patch masking follows the ViT-style patch structure of the base LVLM: the image is divided into $14 \times 14$ patches, and each patch is independently blackened with probability $0.6$.
\end{itemize}

Table~\ref{app:visual_intervention} shows that all intervention strategies outperform the DAPO baseline, confirming the usefulness of counterfactual visual perturbation for improving visual reasoning. 
Among them, random patch masking achieves the best average performance. 
Compared with complete masking, it avoids overly coarse removal of the whole image; compared with Gaussian noise, it introduces more severe visual corruption without severe pixel-level artifacts. 
Therefore, we adopt random patch masking as the default intervention strategy in all main experiments.




\subsection{Visualization of Token Utilities}
\label{app:token_visualization}
\begin{figure*}[htp]
    \centering         
    \includegraphics[width=\textwidth, height=0.4\textwidth]{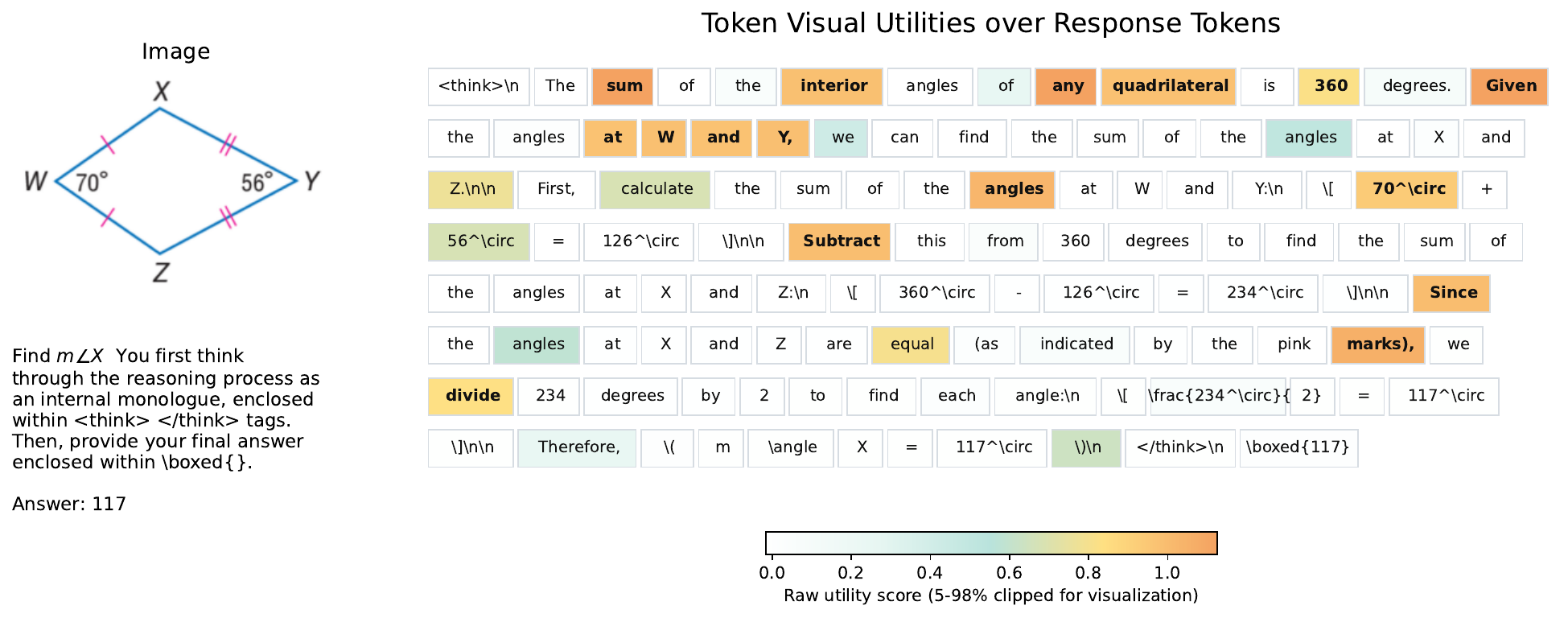}
    \caption{Visualization of token visual utilities. Warmer colors indicate higher utility scores.}
    \label{fig:token_utility_visualization}
\end{figure*}

To better understand how VAA reallocates token-level optimization signals, we visualize the estimated token visual utilities in Figure~\ref{fig:token_utility_visualization}. 
The left side shows the input image and question, while the right side displays the generated response with each token colored according to its visual utility. 
Warmer colors indicate higher utility values and thus stronger contribution to visually grounded reasoning. The visualization shows that VAA assigns higher utilities to tokens that are closely associated with visual perception and key reasoning operations. 
For example, tokens such as ``interior'', ``quadrilateral'', ``angles'', ``70'', ``56'', and ``marks'' correspond to information that must be extracted from the image. 
Meanwhile, reasoning tokens such as ``Given'', ``Since'', and ``divide'' also receive high utility, since they connect the perceived visual facts to the final geometric derivation. 
In contrast, routine function words and formatting tokens generally receive lower utility. 
This suggests that VAA does not uniformly amplify all response tokens, but instead concentrates the RL optimization signal on visually grounded perception tokens and reasoning pivots that are more relevant to solving the problem.

\begin{table*}[htp]
\centering
\setlength{\tabcolsep}{8pt}
\renewcommand{\arraystretch}{1.2}
\definecolor{gainred}{RGB}{230,60,60}
\newcommand{\gain}[1]{\textcolor{gainred}{+#1}}
{\small
\begin{tabular}{lccl}
\toprule
\textbf{Model \& Hardware} & \textbf{Method} & \textbf{Experience Buffer RAM} & \textbf{Avg. Time / Step (s)} \\
\midrule
\multirow{3}{*}{\shortstack[l]{\textbf{Qwen2.5-VL-7B}\\\textbf{(4 GPUs)}}}
    & DAPO                 & --                  & 698.8 \\
    & VPPO                 & --                  & 830.4\textsubscript{\gain{18.8\%}} \\
    & PIVOT                & \textbf{+21.1 MB}   & \textbf{860.8\textsubscript{\gain{23.2\%}}} \\
\bottomrule
\end{tabular}
}
\caption{Computational overhead comparison.}
\label{tab:overhead}
\end{table*}

\subsection{Layer-wise Attention Analysis}
\label{app:attention_analysis}
\begin{figure*}[htp]
    \centering         
    \includegraphics[width=\textwidth, height=0.55\textwidth]{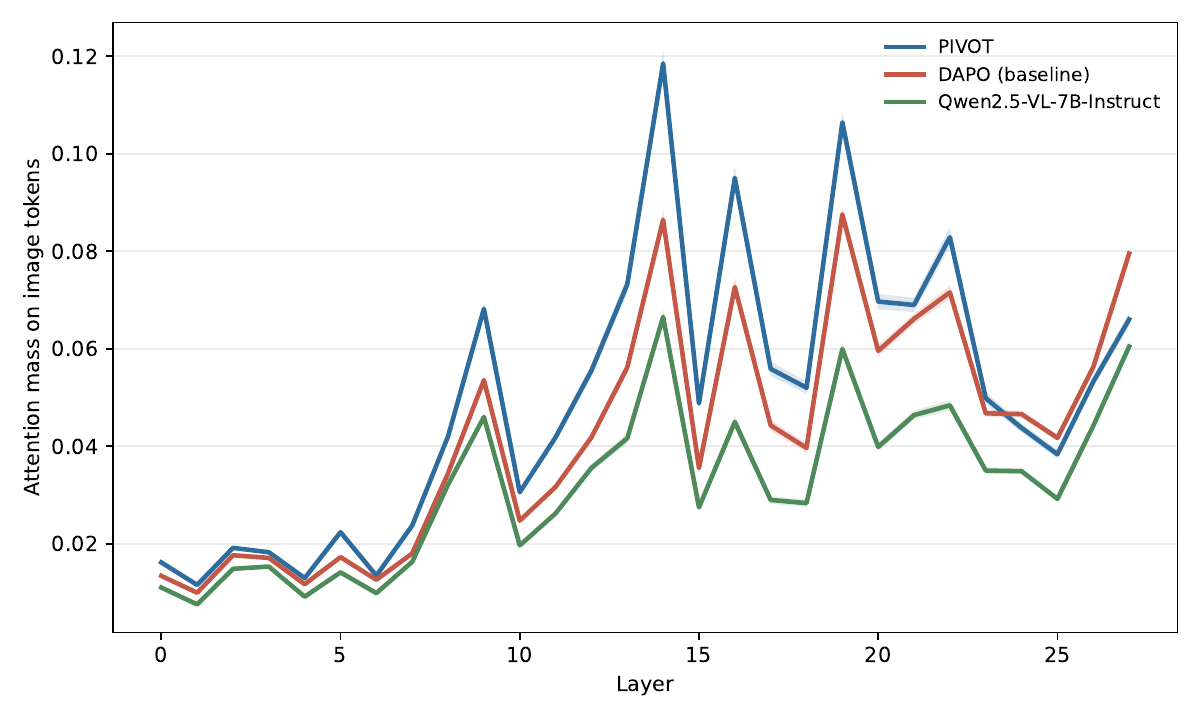}
    \caption{Layer-wise attention analysis. We report the average attention mass assigned by response tokens to image tokens across transformer layers.}
    \label{fig:attn}
\end{figure*}

To further examine whether PIVOT improves the model's reliance on visual information, we conduct a layer-wise attention analysis in Figure~\ref{fig:attn}. 
Specifically, we measure the attention mass assigned by response tokens to image tokens at each transformer layer, and compare PIVOT with the DAPO baseline and the original Qwen2.5-VL-7B-Instruct model.
The results show that PIVOT consistently assigns higher attention mass to image tokens across most layers, especially in the middle and later layers where multimodal reasoning is more actively integrated. 
Compared with the original base model, DAPO already increases visual attention to some extent, suggesting that RLVR encourages the model to exploit visual evidence for solving reasoning tasks. 
PIVOT further strengthens this trend, indicating that trajectory-level anchoring and token-level visual advantage allocation promote stronger visual grounding during response generation. 
Although attention alone cannot fully characterize the reasoning process, this analysis provides complementary evidence that PIVOT encourages the model to attend more to visual tokens when performing multimodal reasoning.

\subsection{Computational Overhead of PIVOT}
\label{app:overhead}
We further analyze the computational overhead introduced by PIVOT in Table~\ref{tab:overhead}. 
All measurements are conducted with Qwen2.5-VL-7B on 4 NVIDIA A100 80GB GPUs. 
Compared with DAPO, VPPO increases the average training time per step from $698.8$s to $830.4$s, mainly due to the additional computation required for visual perturbation-based signal estimation. 
PIVOT further increases the average time per step to $860.8$s, corresponding to a $23.2\%$ overhead over DAPO and only a modest additional overhead over VPPO.
The extra cost of PIVOT mainly comes from two sources: maintaining the buffer for experience replay and estimating token visual utilities for advantage allocation. 
The memory footprint of the experience buffer is lightweight, requiring only $21.1$ MB of additional RAM in our setting (training on Geometry3K dataset). 
Therefore, although PIVOT introduces extra computation during training, its memory overhead remains negligible, and the overall training cost is still practical for LVLM reinforcement learning.

\end{document}